%% file: template.tex
\documentclass{article}

\usepackage{arxiv}

\usepackage[utf8]{inputenc} 
\usepackage[T1]{fontenc}    
\usepackage{hyperref}       
\usepackage{url}            
\usepackage{booktabs}       
\usepackage{amsfonts}       
\usepackage{nicefrac}       
\usepackage{microtype}      
\usepackage{lipsum}
\usepackage{fancyhdr}       
\usepackage{graphicx}       
\graphicspath{{media/}}     

\usepackage{cite}
\usepackage{url,lineno,microtype,subcaption}
\usepackage[onehalfspacing]{setspace}
\usepackage{glossaries}
\usepackage[ruled]{algorithm2e}
\usepackage{listings}
\usepackage{xcolor}
\usepackage{tabularx}   
\usepackage{mathtools} 
\usepackage{amssymb}   
\usepackage{subcaption}                 
\hypersetup{
  colorlinks   = true, 
  urlcolor     = {green!50!black}, 
  linkcolor    = {green!50!black}, 
  citecolor   = {blue!50!black} 
}
\title{\bf Learning to Reason over Scene Graphs:\\ A Case Study of Finetuning GPT-2 into a Robot Language Model for Grounded Task Planning}

\author{
Georgia Chalvatzaki\,$^{1,2,3}$, Ali Younes\,$^{1}$, Daljeet Nandha\,$^{1}$, An Le\,$^{1}$, Leonardo F. R. Ribeiro\,$^{4, \dagger}$ and Iryna Gurevych$^{1,2}$ \\
 $^{1}$ Computer Science Department, Technische Universit\"at Darmstadt, Germany \\
$^{2}$ Hessian.AI, Darmstadt, Germany \\
$^{3}$ Center for Mind, Brain and Behavior, Uni. Marburg and JLU  Giessen, Germany \\
$^4$ Amazon Alexa, Seattle, Washington, United States\\
$\dagger$ Work done while at TU Darmstadt, Germany\\
\texttt{georgia.chalvatzaki@tu-darmstadt.de}\\
}

\date{}

\input{acronyms}

\makeglossaries
\begin{document}
\maketitle
\begin{abstract}
 Long-horizon task planning is essential for the development of intelligent assistive and service robots. In this work, we investigate the applicability of a smaller class of large language models (LLMs), specifically GPT-2, in robotic task planning by learning to decompose tasks into subgoal specifications for a planner to execute sequentially. Our method grounds the input of the LLM on the domain that is represented as a scene graph, enabling it to translate human requests into executable robot plans, thereby learning to reason over long-horizon tasks, as encountered in the ALFRED benchmark. We compare our approach with classical planning and baseline methods to examine the applicability and generalizability of LLM-based planners. Our findings suggest that the knowledge stored in an LLM can be effectively grounded to perform long-horizon task planning, demonstrating the promising potential for the future application of neuro-symbolic planning methods in robotics.
\end{abstract}


\section{Introduction}
The autonomous execution of long-horizon tasks is of utmost importance for future assistive and service robots. An intelligent robot should reason about its surroundings, e.g., regarding the included objects and their spatial-semantic relations, and abstract an action plan for achieving a goal that will purposefully alter the perceived environment. Such an elaborate course of robot actions requires scene understanding, semantic reasoning, and planning over symbols and geometries. The advent of Deep Learning led many researchers to faithfully follow end-to-end approaches due to the representation power of differentiable deep neural networks~\cite{lecun2015deep}. 

The problem of sequential decision-making has been addressed both with search-based and optimization approaches~\cite{kaelbling2011hierarchical,garrett2021integrated,garrett2020pddlstream,toussaint2015logic,driess2019hierarchical}, as well as learning-based~\cite{nair2019hierarchical,hoang2021successor,funk2021learn2assemble} and hybrid methods~\cite{kim2019learning,driess2020deep,ren2021extended,funk2022GRLMIP}. While the first ones enjoy probabilistic completeness, they require full domain specification and have high computational demands. The learning-based methods require broad exploration to learn from experience, but they have shown better generalization capabilities in similar domains to those experienced during training. 

In recent years, \gls{llm} have exhibited an unprecedented generative ability~\cite{bommasani2021opportunities}, thanks to the transformer architecture~\cite{vaswani2017attention} combined with massive datasets distilled from the internet. Naturally, in the quest for general artificial intelligence, researchers try to benchmark such models in reasoning tasks, among others ~\cite{wang2018glue,wang2019superglue}. Robotic embodied intelligence requires both logical and geometric reasoning; hence, it is a holy grail of AI. Several researchers saw a benefit in \gls{llm}, and it was not long before several works started exploring their application to robotics for endowing robots with the long-wished objective of reasoning in the scope of autonomous task planning and interaction~\cite{weichain,brohan2022can}. However, most of these works have focused on the prompting~\cite{brown2020language} and the subsequent prompt engineering~\cite{white2023prompt} 
, in which engineers provide appropriate inputs to \gls{llm} for extracting outputs that can be realizable by a robotic agent, either for human-instruction following~\cite{ouyang2022training} or for planning~\cite{zeng2022socratic,singh2022progprompt}.

This work investigates the applicability of a smaller class of \gls{llm} for robotics, i.e., GPT-2~\cite{radford2021learning}, and we propose and investigate a method that decomposes a long-horizon task into subgoals in the form of a goal-specification for a robotic task planner to execute. The main contribution of our work is the representation of the domain as a scene graph and the subsequent linearization of the graph to keep its structure and relations while forming a suitable input for the finetuning of the GPT-2 model. 
Our thorough experimental evaluation shows that finetuning GPT-2 by additionally grounding its input on the domain can help translate human requests (tasks) to executable robot plans, to learn to reason over long-horizon tasks, as those encountered in the ALFRED benchmark~\cite{shridhar2020alfred}. We compare our proposed approach with classical planning methods to investigate the applicability and generalizability of the \gls{plm}-based planners compared to classical task planners operating on a limited computational budget for a fair comparison. 
We conclude that the knowledge stored in a \gls{plm} can be grounded on different domains to perform long-horizon task planning, showing encouraging results for the future application of neuro-symbolic planning methods in robotics.

\section{State of the Art}
\subsection{Reasoning with Large Language Models} 
\gls{llm} have attracted much attention for understanding the commonsense and reasoning patterns in their latent space~\cite{li2021language,zhou2020evaluating, bian2023chatgpt}.
It has been shown that some abilities in logical and mathematical reasoning seem to emerge when \gls{llm} are prompted appropriately ~\cite{weichain,weiemergent}. However, the engineering effort, as well as the lack of robustness, is a key issue in prompting massive models ~\cite{ruis2022large,valmeekam2022large}. 
While great effort seems to be consumed on few-shot prompting of huge parametric models, it has also been shown by other lines of work show that efficient finetuning of much smaller models~\cite{tay2022transcending}, or the use of small adaptation modules (Adapters)~\cite{houlsby2019parameter,pfeiffer2021adapterfusion} can lead to methods that perform more robustly than large-scale generalist few-shot prompters. In the same direction, the chatbot versions of those huge models raised several points of criticism recently, showing that much more is needed than just prompting a blind human-preference alignment\footnote{\href{https://www.theverge.com/2023/2/9/23592647/ai-search-bing-bard-chatgpt-microsoft-google-problems-challenges}{7 problems facing Bing, Bard, and the future of AI search, from The Verge}}.

\subsection{Robot behavior planning} 
Long-horizon robot behavior planning is an NP-hard problem \cite{wells2019learning}. Current advances in ML and perception led researchers to revisit this fundamental problem, i.e., the execution of multi-stage tasks, whose completion requires many sequential goals to be achieved, considering learning-based heuristics~\cite{driess2020deep}. Researchers consider such problems as \gls{TAMP} problems~\cite{garrett2021integrated,ren2021extended,xu2022accelerating}, with a symbolic plan over entities and predicate with respective action operators with preconditions and effects in the environment. In contrast, a motion plan tries to find a feasible path to the goal. Nevertheless, most \gls{TAMP} methods rely on manually specified rules; they do not integrate perception, and the combinatorial explosion when searching over symbolic and continuous parameters prohibits scaling the methods to challenging, realistic problems~\cite{kim2019learning,garrett2020pddlstream}. 

Transformer models~\cite{vaswani2017attention} that revolutionized the field of \gls{nlp} opened the way for multiple new applications, in particular for robotics, e.g., visual-language instruction following~\cite{pashevich2021episodic}, 3D scene understanding and grounding~\cite{chen2022leveraging,mees2022grounding}, language-based navigation~\cite{huang2022visual,shahlm}. Due to their training on extensive databases, several works explored the use of \gls{llm} for task planning and long-horizon manipulation~\cite{huang2022language}, mainly employing clever prompting~\cite{singh2022progprompt,raman2022planning}, using multimodal information~\cite{zeng2022socratic,jiang2022vima}, grounding with value-functions~\cite{brohan2022can,huanginner2022,chen2022open}, and deploying advances in code generation to extract executable robot plans~\cite{liang2022code}.~\cite{li2022pre} propose to use a \gls{plm} as a scaffold for decision-making policies in interactive environments, demonstrating benefits in the generalization abilities for policy learning even when language is not provided as input or output. 
Recently, PALM-e~\cite{driess2020deep} has integrated a vision transformer with the PALM language model and has encoded some robotic state data to propose a multimodal embodied model, which showed the potential of integrating geometric information of the robot state but achieved limited performance in robotic tasks.

\section{Language models for grounded robot task planning}
\subsection{Problem statement}
Let us assume an agent that is able to move and manipulate objects in an environment, e.g., a mobile manipulator robot in a household environment. Let the environment be composed of a combination of \textit{rooms}, such as `bathroom,' `living room,' or `kitchen.' Each room contains \textit{objects} and \textit{receptacles}, i.e., objects that are able to receive other objects, such as `table,' `drawer,' or `sink.' Each object has (household-specific) properties associated with it that define whether it can be picked up, cleaned, heated, cooled, cut, etc. These properties can change, meaning that the objects have a \textit{state}. The agent can pick up only one object at a time, meaning the agent also has a state, e.g., `object in hand.' Given the fact that the state is preserved over time and future actions depend on past actions, the environment can be characterized as sequential. Therefore, a series  of actions has to be reasoned upon for an agent to be able to execute a series of actions for solving a long-horizon task, i.e., a task that requires the completion of several subtasks and potentially the manipulation of various objects to achieve the end goal.

\subsection{The ALFRED benchmark}\label{sec:dataset}
\begin{figure}[ht]
\vspace{-0.2cm}
    \centering
    \includegraphics[width=0.9\textwidth]{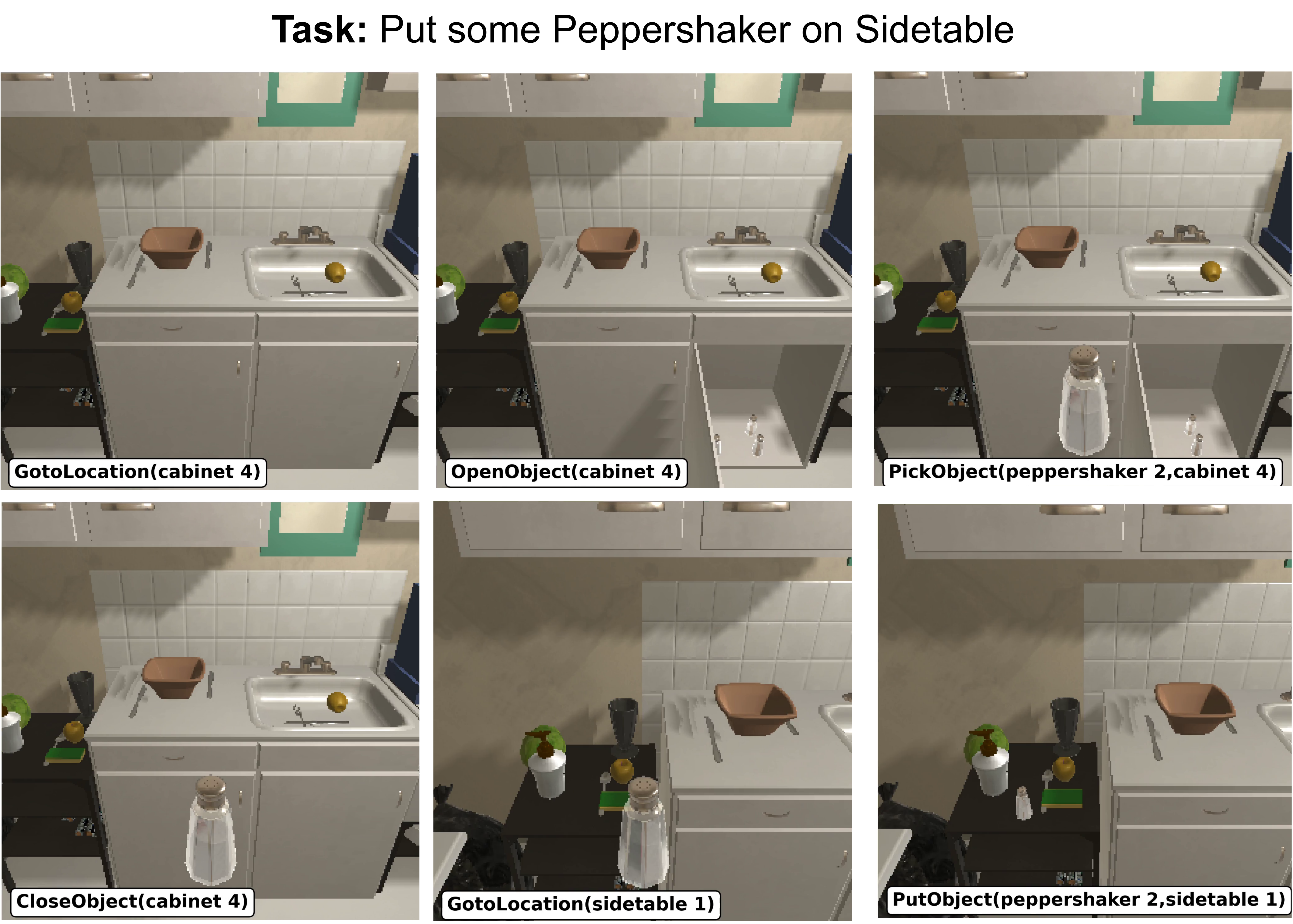}
    \caption{AI2-THOR simulator rendering a sample rollout from the ALFRED~\cite{shridhar2020alfred}. The scenes show a room with household objects and the robot executing a task. Note that the robot does not have an arm, and the object automatically floats in front of the camera; it interacts with the environment through discrete actions. The discrete actions are shown underneath each frame in the form of PDDL commands.}
    \label{fig:ai2thor}
\end{figure}
The ALFRED benchmark~\cite{shridhar2020alfred} contains human-annotated training samples and image-based recordings of everyday household tasks; this is ``25,743 English language directives describing 8,055 expert demonstrations averaging 50 steps each, resulting in 428,322 image-action pair''. In addition to that, the dataset provides a \textit{PDDL domain} of the overall task and a \textit{PDDL problem} for each sample~\cite{aeronautiques1998pddl}. ALFRED heavily depends on AI2-THOR~\cite{kolve2017ai2}, which acts as the underlying controller and simulation environment (based on the Unity game engine): trajectories for each sample of the ALFRED dataset were generated with AI2-THOR and the validation of user-generated actions requires the AI2-THOR controller. Figure \ref{fig:ai2thor} shows a sample scene loaded into the AI2-THOR simulator. Each sample in the dataset consists of a high-level plan in PDDL and the trajectory of the agent's actions which lead to successful task completion, together with a description of the task goal and each plan step in \gls{NL}.

\subsubsection{Data augmentation}
Each sample in the ALFRED dataset can be replayed in the AI2-THOR simulator to collect additional information not contained in the original dataset. ALFRED provides a script that has been modified for that purpose. Data augmentation is necessary for Graph2NL (c.f. §\ref{sec:graph2nl}) to generate a graph representation from the environment state. For each replayed sample, the complete list of objects in the scene, with their respective name, position, and rotation, and the agent position is saved to a separate file next to the trajectory data. This file is later loaded and turned into a processable graph.

\subsection{State and action space}
The state space defines the feedback provided by the environment, while the action space defines the available actions to interact with the environment.

\textbf{State space.}~The AI2-THOR environment can be framed either as fully observable when access to the complete simulator state is given, or partially observable when the agent uses its perception (images) to observe the environment without having access to the true state. AI2-THOR eliminates most physics-related aspects (e.g., objects are automatically picked up and placed by a single action), which makes the highly dynamic and stochastic household environment almost static and deterministic --- almost because some physics still exists. This simplifies the core \gls{TAMP} problem, together with the discrete agent actions coming from the ALFRED dataset. Therefore, the ALFRED benchmark represents an appropriate choice for studying the problem of learning for robotic task planning, where motion failures are minimized by the underlying AI2-THOR controllers. Hence, we can focus on the reasoning aspects of the problem, which is the focus of this study. In this paper, the environment is assumed to be fully observable. The state space, therefore, includes the whole domain as provided by the simulation. In the proposed methods, the state is transformed to \gls{NL} context (§\ref{sec:graph2nl}) and not used directly as a model input.

\textbf{Action space.}~ALFRED has an action space of eight discrete high-level actions: \textit{GotoLocation}, \textit{PickupObject}, \textit{PutObject}, \textit{CoolObject}, \textit{HeatObject}, \textit{CleanObject}, \textit{SliceObject} and \textit{ToggleObject}. The actions \textit{CoolObject}, \textit{HeatObject}, \textit{CleanObject} themselves are a composition of the remaining high-level actions. The underlying AI2-THOR controller also has a discrete action space for the low-level actions: The agent can \textbf{move} \textit{forward}, \textit{backward}, \textit{left} or \textit{right} and \textbf{rotate} \textit{clockwise} or \textit{counter-clockwise} in fixed steps.

\subsection{Task categories}
The ALFRED dataset encompasses seven categories of household tasks: `Look at object', `Pick and place', `Pick two and place', `Pick and place with movable receptacle', `Pick, clean then place', `Pick, cool then place', `Pick, heat then place'. Because, in mobile manipulation, objects can be placed in different corners of a room, each of these tasks includes the sub-problem of navigation.

For the `pick' or `place' subtasks executing the respective \textit{PickupObject} or \textit{PutObject} action is sufficient. But, the subtasks `clean', `cool' and `heat' must be seen as planning problems on their own, because the corresponding actions are a composition of high-level \textit{state-dependent} actions. In regards to the household environment, the subtask `cool' requires a fridge, `heat' requires a microwave (or oven), and `clean' requires a sink as an \textit{receptacle}.

Adding to the difficulty of the `clean,' `cool,' `heat,' and `cut' subtasks is the fact that the ALFRED simulator tracks the state of each object: the subtask is only considered successful when the final object state is correct. For example, if the task category is `Pick, clean then place', the task goal is only completed when the placed object is marked as `clean.'
The implementation aspects of these task categories are discussed in Section \ref{sec:experiments}.

\subsection{RobLM: Robot Language Model for Task plan generation}\label{sec:roblm}

Just like images can be represented by discretizing color space, \gls{NL} can be expressed as a sequence of tokens $\mathbf{x} = [x_1, x_2, ..., x_n]$, where each token is mapped to an embedding (lookup table). The \gls{lm} places a probability distribution $p(\mathbf{x})$ over the output token sequence. $p(\mathbf{x})$ can be decomposed into a conditional probability distribution $p(x_{i+1}|x_i)$, where the probability of each token depends on all previous tokens. This results in the following joint distribution $p(\mathbf{x})$ for a sequence of tokens $\mathbf{x}$:
\begin{equation}\label{eq:1}
p(\mathbf{x}) = \prod_i p(x_i|x_0,x_1,...,x_{i-1})
\end{equation}

In regards to \gls{NN}, $p(\mathbf{x})$ is commonly estimated with the Softmax function~\cite{bengio2000neural} \footnote{The Softmax function applies the exponential function to each element of the input vector and normalizes the values, by dividing by the sum of the exponentials.}
\begin{equation}\label{eq:softmax}
    p(\mathbf{x}) = Softmax(\mathbf{Wh^T + b}) = \frac{\exp(\mathbf{Wh^T + b})}{\sum \exp(\mathbf{Wh^T + b})},
\end{equation}
where $\mathbf{W}$ is the learned weight matrix, $\mathbf{b}$ the bias and $\mathbf{h^T}$ the output vector of the \gls{NN}.
For text generation, the joint probability distribution $p(\mathbf{x})$ (see Equation \ref{eq:1}) can be formulated as a maximum-likelihood objective, where the objective is to maximize the likelihood of the next token occurrence for the given data.

Our goal is to fine-tune a \gls{lm} to get a \gls{roblm} that can generate a complete high-level task plan in one shot, given the domain information and a task goal. Because \gls{lm}s are unsupervised learners, a single training sample contains both \textbf{given} and \textbf{desired} information as \gls{NL} text. A restriction to the text format (a string of characters) comes with challenges: structural information needs to be condensed into a single linear dimension, and conceptually different aspects of the input need to be annotated in the text. This text format, including the syntax, has to be designed in such a way that information can be fed to and extracted from the \gls{lm} reliably.

In \gls{roblm}, the format definition for a \gls{NL} task description must comply with the following syntactic rule (spaces added for readability):
\begin{verbatim}
    Goal [<SEP> Context] <BOS> Plan <EOS>

    [...] := optional
    <SEP> := separator token
    <BOS> := begin-of-sequence token
    <EOS> := end-of-sequence token
\end{verbatim}

\textbf{Goal} is the task goal in \gls{NL}. \textbf{Context} is any additional, yet optional information provided to the \gls{lm}. The task might have ambiguous solutions, and the inherent assumption is that the \gls{lm} will better ``understand'' the task if given a context. Examples of a context are the name of the \textit{room}, the name of the target object, or a \gls{NL} description of the environment (see \ref{sec:graph2nl}).

\textbf{Plan} is the sequence of high-level task actions and their respective arguments, such as objective name or location. Because \gls{plm} have been trained on a diverse corpus of \gls{NL}, including program code, the format for plans follows syntactical rules similar to that of a generic programming language:
\begin{verbatim}
    Action0(arg0[,arg1]); Action1(arg0[,arg1]); ...
\end{verbatim}
The sequence between the special tokens $<\text{BOS}>$ and $<\text{EOS}>$ can be extracted to retrieve the plan from the \gls{lm}-generated output.

\subsection{Training}
\begin{figure}[ht]
    \centering
    \includegraphics[width=0.75\textwidth]{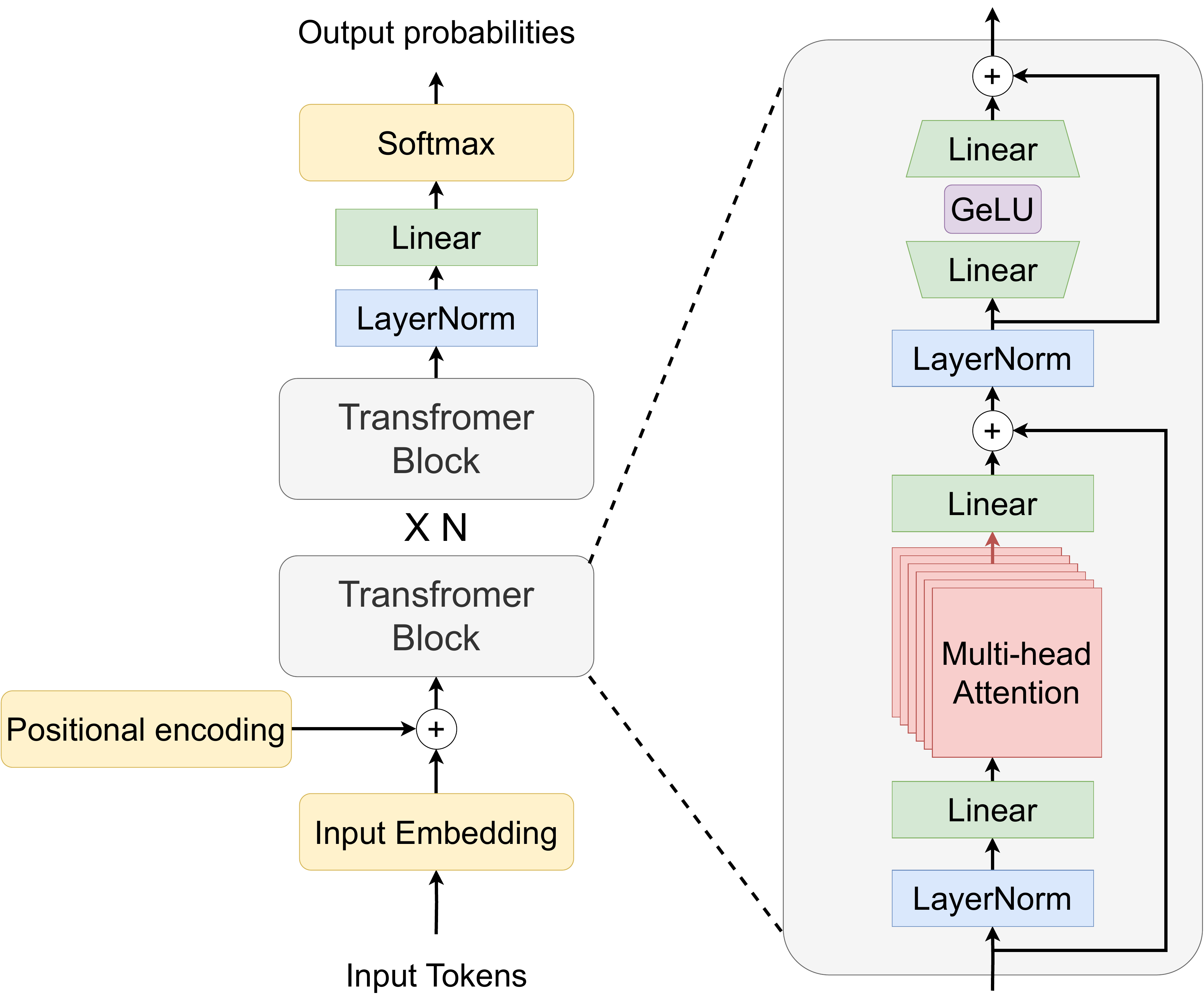}
    \caption{Decoder-only Transformer architecture. The input to the decoder is tokenized text, and the output is probabilities over the tokens in the tokenizer vocabulary. The positional encoding is added to the embedded input to account for the order. Transformer's decoder can have multiple transformer blocks, each of which contains multi-head attention with linear layers and layer normalization. GPT models uses GeLU activation between the last linear layers. }
    \label{fig:transformer}
    \vspace{-0.5cm}
\end{figure}

\gls{roblm} generate a plan as a text given the goal and the context, which involves causal language modeling for text generation. Decoder-only autoregressive language models \ref{fig:transformer} are frequently used for the problem of text generation. Hence, we chose GPT2 model as the base model for \gls{roblm}. 
\gls{roblm} uses the base version of the GPT-2 \gls{plm} (`gpt-2')~\cite{radford2021learning}, loaded and initialized with pre-trained weights from the Huggingface \cite{wolf2019huggingface} Transformer library. 
Fine-tuning GPT-2 for causal language generation has a self-supervised setup, where the labels are the inputs shifted to the right, which entitles learning to predict the next token in a sequence.

The GPT-2 \gls{plm} is fine-tuned to the pre-processed training data of the ALFRED dataset, which has around 20.000 samples, with three sets of \gls{NL} descriptions for each sample. The ADAM \cite{kingma2014adam} optimizer is used with a learning rate of $5e^{-5}$ and the \gls{lm} is trained for two epochs. Fine-tuning a GPT-2 \gls{lm} to the ALFRED training data with a single GPU-accelerated computer takes around 30 minutes (27 iterations/s - measurement not representative due to hardware dependence).

\subsection{Generation pipeline}
\begin{figure}[ht]
    \centering
    \includegraphics[width=0.9\textwidth]{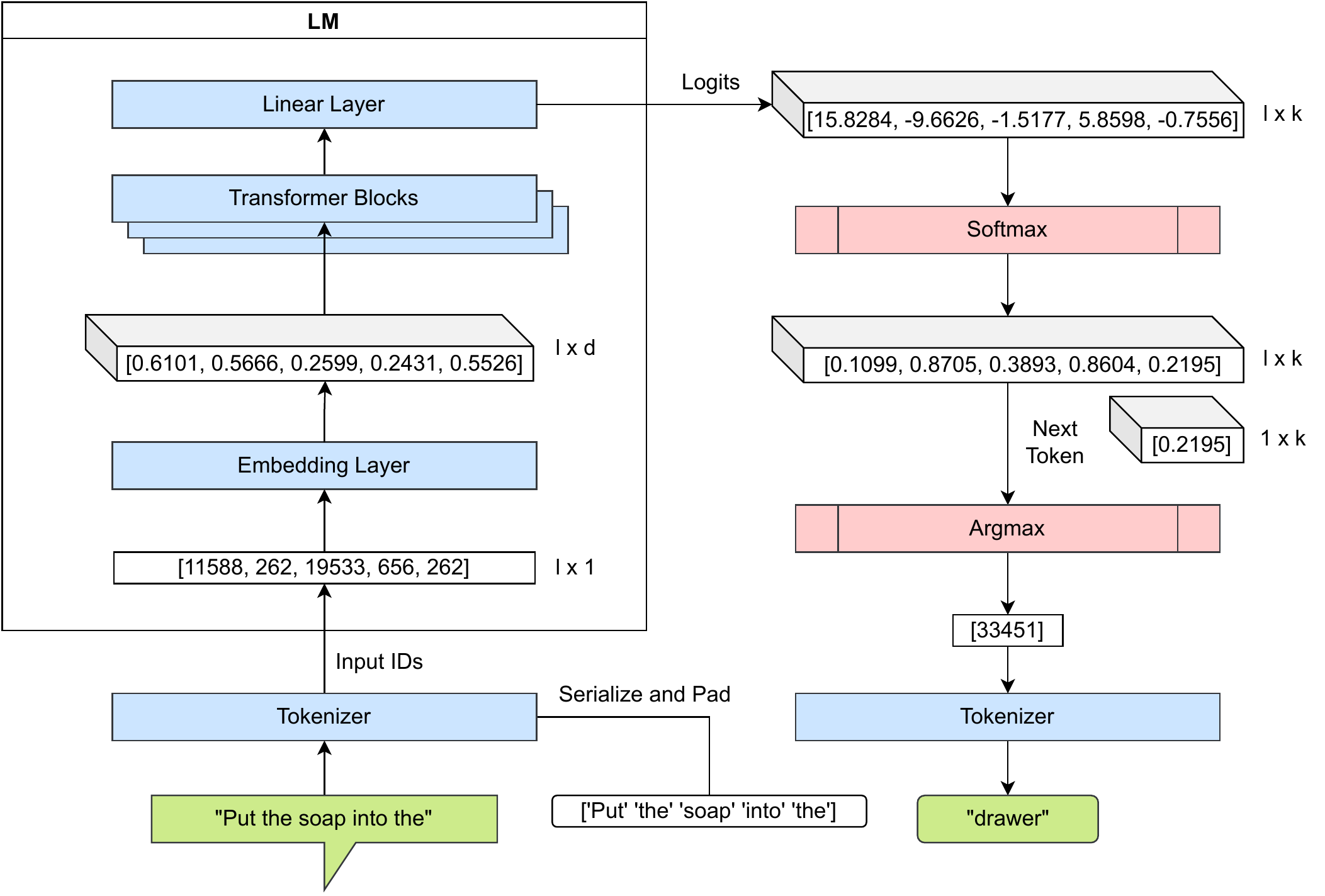}
    \caption{Illustration of a forward pass through \gls{roblm} for text generation with a greedy next-token selection strategy. The forward passes are repeated in a recursive manner until an end-of-text token is encountered or the defined sequence limit is reached.}
    \label{fig:pipeline}
    \vspace{-0.5cm}
\end{figure}

For inference, RobLM takes only the \gls{NL} task goal together with an optional context and outputs the complete step-by-step plan for completing the goal. This plan is composed of high-level instructions rather than low-level controller commands.

\noindent\textbf{Example}~Given the task ``Put the soap into the drawer:'', \gls{roblm} (no context) generates the following plan:
\begin{verbatim}
    Put the soap into the drawer:
    0.GotoLocation(countertop)
    1.PickupObject(soap)
    2.GotoLocation(drawer)
    3.PutObject(soap,drawer)
\end{verbatim}

The plan is generated by consecutive forward passes through the Transformer model. For a vocabulary size of $k$ and a token sequence of length $l$ (with $l \leq 1024$ for GPT-2), the forward pass of the Transformer yields an output vector of size $k \times l$ with values in the interval $[0, 1]$.

The Transformer outputs \textit{scores}, called \textit{logits}, for each token in the input sequence. These scores are converted to a probability distribution $p(\mathbf{x})$ by using the Softmax function, as described in Equation \ref{eq:softmax}.

Two possible generation strategies for selecting the next token from $p(\mathbf{x})$ are: \textit{greedy search} and \textit{top-k / top-p sampling} ~\cite{holtzmancurious}. In the greedy strategy, the token $x_{sel}$ with the highest likelihood is picked with $x_{sel} = \arg \max {p(\mathbf{x}})$. In the top-k sampling strategy, as the name suggests, the scores are sorted, and one of the first \textit{k} candidate tokens is randomly sampled. By extending the top-k sampling with an additional top-p strategy, the sum of the $k$ candidates must be equal to or greater than $p \in [0, 1]$. Simply put, top-k widens the choice over the next tokens and top-p filters out low-probability tokens.

Figure \ref{fig:pipeline} illustrates, on the basis of an example, a forward pass through the Transformer with a greedy selection strategy. These steps are repeated recursively until an end-of-text token is encountered or the defined sequence limit is reached to generate the full plan.

This \gls{lm} model was fine-tuned to generate a structured output, omitting special tokens, characterized by numbered actions and their arguments in parenthesis. Note that the input is always part of the output, due to the generation function utilized by \gls{roblm}.

Note that it is not guaranteed that the `soap' can be found inside the `drawer' on the `countertop'. In fact, it could be in any possible location permitted by the environment. However, given a greedy search strategy, for the given task goal the \textit{likelihood} for the `soap' being on the `countertop' is the highest in this case.

\subsection{Transforming Scene Graphs to Natural Language: Graph2NL}\label{sec:graph2nl}
\gls{plm}s are trained on \gls{NL}. Because of this, \gls{NL} is a natural modality for fine-tuning a \gls{plm}. When a context is provided to the \gls{lm}, this context must be presented in \gls{NL} just like the input sequence. If the context should encapsulate the environment state, this means that the state has to be transformed into \gls{NL} before being supplied to the \gls{plm}.

Graph2NL is a novel method that ``translates'' the object-centric scene graph representation of the environment state to \gls{NL}. Optionally, domain knowledge about the environment\footnote{Domain knowledge entails every possible room, object, and receptacle name and their allowed relations, as described in the respective documentation: https://ai2thor.allenai.org/ithor/documentation/objects/object-types} can be infused into this graph. The following steps describe the core Graph2NL process:
\begin{enumerate}
    \item Generate an object-scene graph $G$ with a node for the agent and nodes corresponding to objects, node attributes being the position and rotation of the object in Euclidean space and their respective distance and orientation vectors as edge attributes.
    \item (Optional) Infuse domain knowledge about the environment by connecting all dependent nodes and all nodes reachable by the agent.
    \item Connect the agent (node) to all reachable nodes, if given domain knowledge, or to all nodes, if not given domain knowledge.
    \item Given a task and the identified target object, find all paths in the graphs leading from the agent (node) to the target object (node).
    \item Use edge attributes in the found paths to describe the task-centric environment state, by mapping geometric relations to \gls{NL} tokens.
\end{enumerate}

\subsubsection{NL mapping}

To translate geometric relations attributed by the graph edges into a \gls{NL} description, a mapping function is designed. In human speech, distances are expressed by a vocabulary of words such as ``close'' or ``far'' and orientations are expressed by words such as ``in front'' or ``behind''. Graph2NL adapts this vocabulary to describe the (numeric) distance and orientation from one node relative to another in \gls{NL}.
\begin{table}[ht]
    \caption{Graph2NL mapping table. Distances are mapped to \gls{NL} vocabulary (or a symbol) in a one-to-one relation. \textit{Yaw} describes the orientation along the surface normal when viewed from a top-down perspective, and \textit{Pitch} describes the z-planar offset (altitude) in relation to the origin.}
    \label{tab:mapping}
    \centering
    \begin{tabular}{c|c|c}
        \hline
        \multicolumn{3}{c}{Distance [m]} \\
        \hline
        Value & \gls{NL} & Symbol \\
        \hline
        $> 5$ & distant & a \\
        $> 4$ & far & b \\
        $> 3$ & reachable & c\\
        $> 2$ & near & d \\
        $> 1$ & close & e \\
        $> 0.5$ & closer & f \\
        $> 0.1$ & next & g \\
        $< 0.1$ & in & h \\
        \hline
    \end{tabular}
    \hspace{10pt}
    \begin{tabular}{c|c|c}
        \hline
        \multicolumn{3}{c}{Yaw [\textdegree]} \\
        \hline
        Value & \gls{NL} & Symbol \\
        \hline
        45 to 135 & right & i \\
        135 to 225 & back & j \\
        225 to 315 & left & k  \\
        315 to 45 & front & l \\
        \hline
        \multicolumn{3}{c}{Pitch [\textdegree]} \\
        \hline
        Value & \gls{NL} & Symbol \\
        \hline
        $\geq 0$ & above & m \\
        $< 0$ & below & n \\
        \hline
    \end{tabular}

\end{table}

Table \ref{tab:mapping} summarizes the mapping used in Graph2NL. The distance between nodes is expressed in Cartesian space and orientation in polar coordinates, where \textit{Yaw} is the azimuth angle (rotation along the surface normal) and \textit{Pitch} is the zenith angle (altitude). With this mapping, the geometric relation between two nodes can be explained by three words (one for each: distance, pitch, and yaw).

The vocabulary contains 8 words to express the distance, 4 words to express the vertical, and 2 words to express the horizontal orientation. Combinatorially, this gives 64 possible geometric configurations. By treating each of these configurations as a relation and assigning a special symbol (token) for each relation, the geometric relationship is expressed in a condensed form. A simple approach, referring to the \textit{Symbol} column in Table \ref{tab:mapping}, is by assigning a symbol to each word. By combining the symbols for distance, pitch, and yaw, the condensed (three-letter) representation of the geometric relationship is created.

These symbolic representations can optionally be added to the \gls{lm} tokenizer as so-called \textit{special} tokens. Shorter token sequences, in general, decrease both the training and inference time.

\begin{figure}[ht]
    \centering
    \includegraphics[width=.8\textwidth]{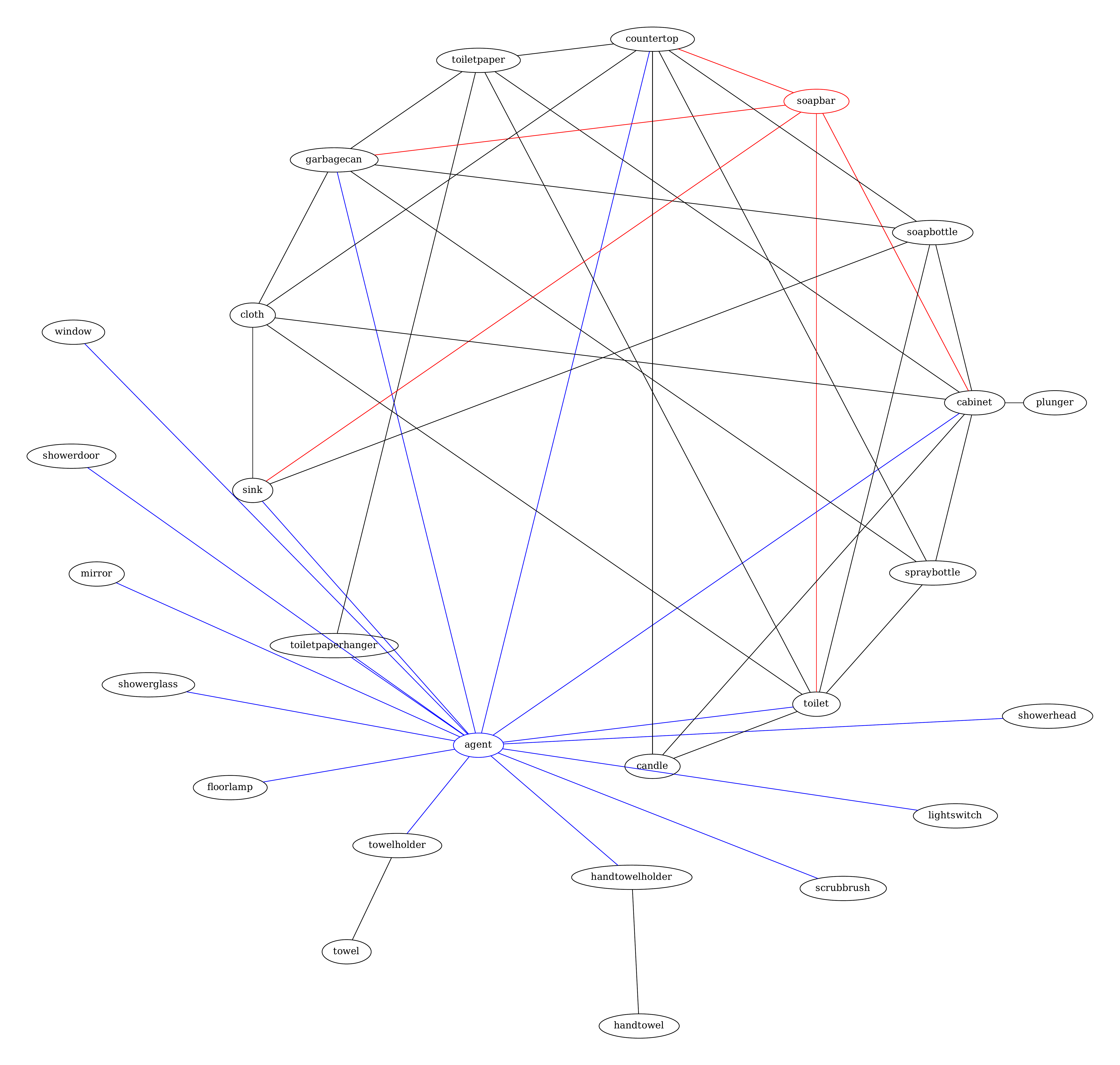}
    \caption{Graph2NL example graph. After locating the root (`agent') and target node (`soapbar'), the shortest paths connecting those nodes are found and summarized in \gls{NL} by mapping all edge attributes along the path.}
    \label{fig:graph}
\end{figure}

\textbf{Example}~Let the task goal be: ``Put the soap into the drawer''. The input query to Graph2NL would consist of the target object `soap'. Figure \ref{fig:graph} shows the graph constructed by Graph2NL from the augmented data (§\ref{sec:dataset}), including domain-specific knowledge. 
After finding the shortest paths between the root (`agent') and target node (`soapbar'), Graph2NL produces an output in the following form (cut-off at search depth 2):
\begin{verbatim}
    [Bathroom=
    - closer below left sink near below back soapbar
    - closer below left cabinet near above back soapbar
    - closer above left countertop next above back soapbar
    - close below back toilet closer below back soapbar
    - closer below back garbagecan close below back soapbar]
\end{verbatim}
The \gls{NL} context by Graph2NL starts with the name of the room extracted from the scene graph, followed by the geometric description of each node connected to the target node on the path from the agent. `-' indicates the root note, which is the agent.

For this example, Graph2NL would produce the following \textbf{condensed} form:
\begin{verbatim}
    [Bathroom=
    - fnk sink dnj soapbar
    - fnj cabinet dmj soapbar
    - fmk countertop gmj soapbar
    - enj toilet fnj soapbar
    - fnj garbagecan enj soapbar]
\end{verbatim}
This form of state representation is unique for each problem configuration and forms the context that grounds \gls{roblm}.

\section{Experiments}\label{sec:experiments}

\subsection{Preliminary analysis for task plan generation with GPT-2 and GPT-3}
 \gls{llm}s can represent knowledge from the data they have been trained on, however, the question remains, if this knowledge can be leveraged to solve planning tasks, i.e., can \gls{lm}s reason?
 This is investigated by comparing the text-generation results of GPT-2 (\gls{ZSL}) and GPT-3 (\gls{fsl}) for a planning task.

Let an example planning task be the following:
\begin{verbatim}
    Given an instruction to a household robot,
    formulate the steps to complete the instruction.
    
    The instruction is: ``Put a washed slice of apple on the table.''
\end{verbatim}
The results for this task vary greatly between the two \gls{lm}s.

\textbf{GPT-2} (`gpt-2' model, 1.5B parameters) completely fails this task and produces an output that resembles a repetition of phrases in the input sentence:
\begin{verbatim}
    Put a washed slice of apple on the table.
    Put a washed slice of apple on the table.
    Put a washed slice of apple on the table.
    ...
\end{verbatim}
Similar behavior has been observed for other \gls{lm} falling into the \gls{ZSL} category. The input sentence is not an open-ended question and requires a high level of reasoning capabilities.

\textbf{GPT-3} (`text-davinci-002' model, 175B parameters), when given the same instruction as input, is able to make assumptions and formulate a whole plan based on these assumptions:
\begin{verbatim}
    Assuming the apple is not cut and not washed.
    
    1. Pick up the apple from the counter.
    2. Cut the apple into a slice.
    3. Wash the apple slice.
    4. Place the apple slice on the table.
\end{verbatim}
The \gls{fsl} paradigm allows GPT-3 to be very sensitive to context changes and seemingly \textit{understand} the request at hand. However, smaller GPT-3 \gls{plm}s (GPT-3 curie, GPT-3 babbage, GPT3-ada) show a degraded quality in the produced plan. \cite{floridi2020gpt} have shown that GPT-3 would not pass the Turing test\footnote{a test that makes an artificial system indistinguishable from a human}, as to having ``no understanding of the semantics and contexts of the request, but only a syntactic (statistical) capacity to associate words [...]''.

These tests have shown that plan generation capabilities of \gls{llm}s vary dramatically depending on the underlying learning paradigm, model architecture, and parameter size. GPT-2, out of the box, is completely unsuited for solving planning tasks that require a minimum level of text understanding. However, as later (§\ref{sec:roblm}) shown, GPT-2 is able to successfully generate plans when fine-tuned to a training dataset (§\ref{sec:dataset}). The question of whether a fine-tuned GPT-2 model can \textit{leverage} knowledge for planning is addressed in the following section. GPT-3, unfortunately, is only accessible through a paid service by OpenAI as part of the OpenAI Beta program\footnote{OpenAI Beta program: https://beta.openai.com/playground} (requires registration). Fine-tuning of own GPT-3 models is possible through the provided service. Practical applications, however, are limited because each query has to be sent to and processed by the OpenAI service. Even if a \gls{plm} was made available, the hardware requirements for running GPT-3 models are immense, even for today's standards, due to the sheer parameter count. It is for these reasons that GPT-3 is not considered as a basis for fine-tuning to \gls{roblm}.

\subsection{Evaluation of RobLM}
This section presents the main experiments conducted for evaluation of \gls{roblm}. We first define the appropriate metrics and a baseline method required to make the evaluations measurable and comparable. The \textit{grounding} problem is explained in accordance with the practical aspects of integrating the available methods into the simulator. For the experimentation part, a set of fine-tuned \gls{llm}s is compared with the baseline performance. 

\subsubsection{Metrics}
To validate a fine-tuned \gls{lm}, only the \gls{NL} task goal of each validation sample and optionally, the context is fed to the RobLM generation pipeline (see Figure \ref{fig:pipeline}). Validation is performed over each task category rather than all the validation data. This enables the analysis of a task-dependent performance: some task categories are more complex than others leading to a longer trajectory of actions and hence an increased difficulty.

Two metrics are defined for validation: \gls{lm} accuracy and plan success rate.

\textbf{Definition --- Accuracy.}
Accuracy measures how accurately the \gls{lm} is able to predict the following parts of the plan:
\begin{itemize}
    \item the correct count and names of all actions in the plan (action accuracy)
    \item the correct count and names of all arguments in the plan (argument accuracy)
    \item the correct count and names of all actions \textbf{and} arguments in the plan (``full plan'' accuracy)
\end{itemize}
For a found plan, the accuracy of actions and arguments counts if \textbf{all} actions or arguments are correct. With this metric, it is possible to anchor the cause of plan failure to either the actions or the arguments, or both.

Having an accurate \gls{lm} does not necessarily mean that the generated plan leads to success --- at least, as long the ``full plan'' accuracy is below $1.0$, i.e., the trajectory is not replicated perfectly. A second metric is required that measures the actual \textit{success rate} of the fine-tuned \gls{lm} in simulation. There are two possible scenarios that justify this additional metric: First, the plan could fail in simulation, even if it seems accurate. And second, the plan could succeed in simulation, even if the plan is not completely accurate.

\textbf{Definition --- Success rate.}~The success rate is a measure of the successful completion of individual sub-tasks of a validation task. After loading the trajectory, environment state, and goal from the validation sample into the AI2-THOR simulator, the actions predicted by the \gls{lm} are translated into low-level controller actions via task and geometric grounding (§\ref{sec:grounding}), which are then passed to the AI2-THOR controller and executed in the simulator. After every simulator step, a check is performed to determine whether the target conditions for sub-task completion have been met. If the target conditions are kept unsatisfied after execution of the last low-level action, it counts as a success towards the sub-task, or otherwise, as a failure.

\subsubsection{Baseline}
A baseline is an oracle, or upper bound, that serves as a measurement reference. \gls{fd}~\cite{helmert2006fast} is used as the baseline for evaluation. We consider a classical task planner like \gls{fd} appropriate since it also has access to the full domain and is a complete algorithm~\cite{helmert2006fast}. Therefore, the ability of a \gls{roblm} to match or outperform \gls{fd} (for a given time budget) would reveal whether \gls{lm}s can be helpful towards learning task planning.
Every ALFRED validation sample comes with a PDDL problem file, while the PDDL domain is shared by all tasks; this allows the PDDL planner to generate a plan for each sample.

To generate a plan using \gls{fd}, the PDDL problem files provided by ALFRED have to be pre-processed. \gls{fd} is able to handle \gls{adl} instructions, as found in the PDDL problem, but is not able to process optimization-related additional information present in the files.

\subsubsection{Hardware setup}
For fine-tuning \gls{lm}s and evaluating each model, we used the Lichtenberg Cluster of TU Darmstadt, which contains stacks of NVIDIA® A100 and V100 GPUs. Internal tests have shown that a single GPU can decrease the training time by a factor of 10 (these tests are not representative because performance depends on every hardware component). To run experiments in the AI2-THOR simulation, we used a PC with an NVIDIA® RTX 3080Ti GPU.

\subsubsection{Instruction grounding}\label{sec:grounding}
\textit{Grounding} van be defined as mapping a high-level, abstract, or symbolic representation to a low-level (ground) representation. Grounding of an abstract plan to objects is called object or geometric grounding (or ``world grounding''), and grounding of \gls{NL} to robot tasks is called task grounding. In this case, instructions generated by the \gls{lm} are made up of actions that require a \textit{task grounding}, and arguments, which require a \textit{geometric grounding}.

\subsubsection{Task grounding}
Plans generated by RobLM consist of high-level actions and are not directly executable by the AI2-THOR controller. Each possible action predicted by the \gls{lm} has to be grounded to a task, which then translates to a sequence of low-level controller actions.

For task grounding, three possible types of tasks are defined: navigation, manipulation and composite. In a navigation task, the agent is required to move from one to another location. In a manipulation task, the agent performs an action affecting the environment state. Composite tasks are a composition of manipulation tasks that need to be completed in a specific order.

Task grounding is performed as follows:
\begin{itemize}
    \item The action \textit{GotoLocation} is grounded to the navigation task and delegated to a trajectory planner for navigation (see below).
    \item The actions \textit{PickupObject}, \textit{PutObject}, \textit{ToggleObject} and \textit{SliceObject} are grounded to the manipulation task, the actions can be directly executed by the low-level controller.
    \item The actions \textit{HeatObject}, \textit{CoolObject} and \textit{CleanObject} are grounded to the composite task, which is translated to this sequence of low-level actions: \textit{ToggleObject} $\rightarrow$ \textit{PutObject} $\rightarrow$ \textit{ToggleObject} $\rightarrow$ \textit{ToggleObject} $\rightarrow$ \textit{PickupObject} $\rightarrow$ \textit{ToggleObject} (example given below).
\end{itemize}

For example, the action \textit{HeatObject} in a scene containing a microwave would be grounded to the composite task and translate to a series of low-level actions that involve opening the microwave (\textit{ToggleObject}), putting the target object inside (\textit{PutObject}), closing the microwave (\textit{ToggleObject}), waiting (no action), opening the microwave (\textit{ToggleObject}), picking up the heated target object (\textit{PickupObject}) and finally closing the microwave (\textit{ToggleObject}).

\subsubsection{Geometric grounding}
An argument can be anything from a location to an object name. An argument produced by the \gls{lm} might be ambiguous or non-existing in the environment. In order to be understood by the controller, these arguments have to be grounded on a geometric level.

For grounding arguments, first, all available objects are retrieved from the simulation. Then, the world coordinates of all objects matching the predicted symbol (target object) are gathered. E.g., if the predicted target object is `soap', the position of all `soap'-type objects can be queried and retrieved from the simulator. The low-level control commands are finally generated with the help of the ground-truth navigation graph of the scene.

\subsubsection{Navigation}
By overlaying the world with a grid, every position in the world is given a discrete coordinate. A navigation graph (not to be confused with a \textit{scene graph} or Graph2NL graph) creates a node for each coordinate and connects all the nodes that are \textit{accessible} one from another.

Similar to the procedure of Graph2NL (§\ref{sec:graph2nl}), the navigation graph is traversed after locating the agent and target node by the object name. A search algorithm is used to find the shortest path in the graph from the agent to the target object - in this case, it is the A* algorithm\footnote{Refer to \cite{duchovn2014path} as a work on path planning in mobile robotics using the A* search algorithm}. The search returns a sequence of nodes, which corresponds to a sequence of coordinates (a trajectory). Lastly, a motion planner takes the trajectory as an input and outputs a sequence of low-level controller actions (AI2-THOR conveniently provides a motion planner for navigation).

\subsubsection{Experimental results}\label{sec:exp_roblm}
A set of fine-tuned \textbf{RobLM} models are evaluated against the baseline. The fine-tuned models differ in the amount of context provided during training time:
\begin{enumerate}
    \item `No context' --- Only task goal
    \item `Scene knowledge' --- List of all available objects in the environment, found in the PDDL problem
    \item `Scene graph' --- Description of geometric relations to the target object, generated by Graph2NL
    \item `Full context' --- Description of geometric relations of all objects, generated by Graph2NL
\end{enumerate}
Given a PDDL problem file, Graph2NL automatically generates the context in the specified text format. This context is provided to the \gls{lm} for training and inference.

\paragraph{Results --- Accuracy}
\begin{figure}[ht]
    \centering
    \includegraphics[scale=0.7]{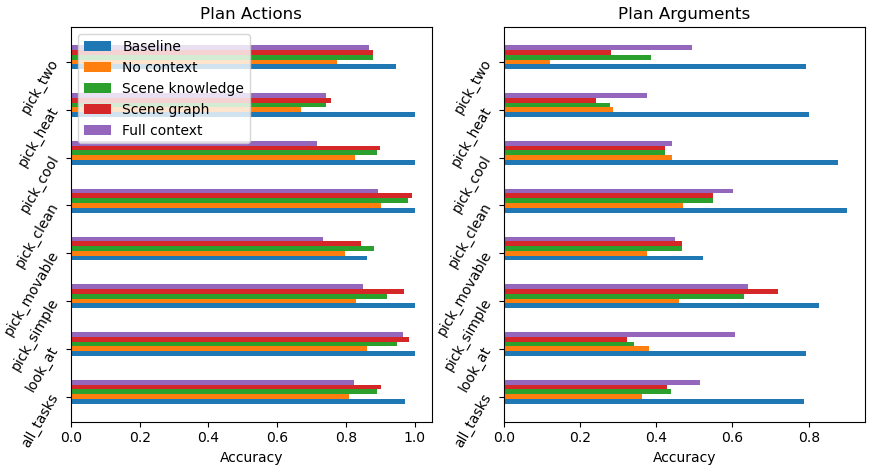}
    \caption{Prediction accuracy of actions and arguments for previously \textbf{unseen} data across a set of tasks. Neither \gls{roblm} model is able to outperform the baseline (blue) but shows high accuracy in the prediction of plan actions. Context-driven models (green, red, and purple) perform better than the model without any scene-related context (orange). 
    }
    \label{fig:accuracy}    
\end{figure}

Figure \ref{fig:accuracy} summarizes the evaluation of the fine-tuned RobLM models compared to the \gls{fd} baseline for previously unseen (validation) data. It can be observed that none of the fine-tuned \gls{roblm} models is able to outperform the baseline. Going through each of the models and starting with the `No context' model it is surprising that this model, even without any contextual information, is able to generate the correct plan actions with high accuracy. The `Scene knowledge' and `Scene graph' models have a similar performance, the `Scene graph' generally being slightly more accurate in both actions and argument prediction. Both of these models overall outperform the `No context' model, with a significant improvement of the context models in the arguments prediction.

Given these results, the following conclusions about the examined models can be made:
\begin{enumerate}
    \item Failed plans are mostly caused by wrong arguments (objects or locations) and only in some cases by wrong actions.
    \item The \gls{lm} is able to learn the structure of tasks, but not scene-dependent components.
\end{enumerate}

\gls{roblm} is able to distinguish between the task categories and provide a correct task action plan. However, where this model fails is in finding all correct action arguments, i.e., locations and object names. This can be explained by the fact that the task goal alone does not reveal the actual location of the target object. Because the target can be in any accessible location in the environment, or in any accessible receptacle, the produced argument is the result of the \gls{lm} imitating the \textit{most-likely} cases observed in the training data.

Overall, these results are consistent with the point made on contextual information and prediction accuracy: giving the model information about the environment, i.e., fine-tuning a model to be grounded to the scene, \textbf{does} improve performance.

\subsection{Results --- Success rate}
\begin{table}
    \caption{Success rates of sub-task completion in simulation --- RobLM (`No context') compared to the baseline on seen and unseen validation data.}
    \label{tab:successrate}
    \centering
    \begin{tabular}{c|c|c|c|c|c}
    \hline
        \textit{Success rate} & \multicolumn{2}{c}{Baseline} & ~ & \multicolumn{2}{c}{RobLM (`No context')} \\
        \hline
        Task & seen & unseen & ~ & seen & unseen \\
        \hline
        GotoLocation & 0.318 & 0.393 & ~ & \textbf{0.422} & \textbf{0.499} \\
        PickupObject & 0.466 & 0.474 & ~ & \textbf{0.776} & \textbf{0.749} \\
        PutObject & \textbf{0.385} & \textbf{0.331} & ~ & 0.116 & 0.092 \\
        SliceObject & 0.629 & 0.5 & ~ & \textbf{0.94} & \textbf{0.98} \\
        ToggleObject & 0 & 0 & ~ & \textbf{0.84} & \textbf{0.864} \\
        \hline
    \end{tabular}

\end{table}

A plan is successful if each of the sub-tasks for a stated task is completed. Table \ref{tab:successrate} summarizes the success rate of \gls{roblm} compared to the baseline for actions of the navigation task (\textit{GotoLocation}) and manipulation task (\textit{PickupObject}, \textit{PutObject}, etc.). Composite tasks have been omitted from this evaluation because of high failure rates caused by their task grounding complexity.

Regarding geometric grounding, arguments predicted by \gls{roblm} are grounded to all matching objects in the world, and \gls{roblm} is allowed to ``try'' all possibilities. E.g., when the objective is to ``Get soap'', multiple `soap'-type objects could exist in the scene. Each possibility given by the geometric grounding is simulated by storing and restoring the simulator state. While this is a clear advantage for \gls{roblm} over the baseline, the evaluation still holds because the \gls{lm} is required to predict the correct location or object names.

Based on the presented results, the \gls{lm}-based system performs exceptionally well on sub-tasks requiring the action \textit{PickupObject}, while the action \textit{PutObject} does not succeed equally well, since it does not come close to the baseline performance.

Overall, the success rate of the baseline method is not nearly as high as expected, hinting at potential implementation-specific failures in the task grounding and in the low-level controller interaction with objects. In the low-level controller, visual information is not included. This means that the robot is controlled in a ``blind flight'' mode. The AI2-THOR simulation requires the target object to be in \textit{view}. If the object is not visible, e.g., because the agent is looking in the wrong direction, the interaction fails and with it, the sub-task. Because of the fact that both systems have been evaluated within the same framework, these results \textbf{do not} dismiss a potential use-case for \gls{lm} in planning.

\subsubsection{Additional Results}
We provide additional experiments  for a deeper analysis of potential points of failure of RobLM. These experiments entail a different sampling strategy and context refinement.

\paragraph{Top-k / top-p sampling}
\begin{table}[ht]
    \caption{Top-k and top-p sampling ($k=10$ and $p=0.9$) --- tokens are sampled three times for the `Pick Simple' task, giving only slight deviations in the final accuracy.}
    \label{tab:topk}
    \centering
    \begin{tabular}{c|c|c|c}
    \hline
    \multicolumn{4}{c}{RobLM `No context' model, `Pick Simple' task} \\
    \hline
    Accuracy of & 1st sample & 2nd sample & 3rd sample \\
    \hline
    Actions             & 0.7746 & 0.8169 & 0.7817  \\
    Arguments           & 0.3803 & 0.4085 & 0.3944  \\
    \hline
    GotoLocation       & 0.8772 & 0.9123 & 0.8904  \\
    PickupObject       & 0.8380 & 0.8662 & 0.8451  \\
    PutObject          & 0.7971 & 0.8227 & 0.7986  \\
    \hline
    GotoLocation\_Args & 0.5658 & 0.5877 & 0.5833  \\
    PickupObject\_Args & 0.7535 & 0.8028 & 0.7746  \\
    PutObject\_Args    & 0.6449 & 0.6667 & 0.6331 \\
    \hline
    \end{tabular}

\end{table}
So far, every experiment conducted has used a greedy next-token selection strategy. In order to be able to tell with certainty that a found plan is the ``best possible'' plan, a comparison with another sampling strategy is required. This additional experiment repeats the previous one, but this time with a top-k and top-p sampling strategy. The comparison is done with the `No context' RobLM model, for all tasks with $k=10$ and $p=0.9$, i.e., tokens are sampled from the top-10 predictions and sum up to a probability $\geq 0.9$.

Since a similar pattern was observed in the individual task evaluations, Table \ref{tab:topk} reports the results for the `Pick Simple' task only. Each token is sampled three times, giving three possible solutions to be evaluated. Slight --- uniform and hence dismissable --- variations in the prediction accuracy exist between these three runs. The sampling-based method performs slightly worse than the greedy strategy.

\paragraph{Refined context}
\begin{figure}[ht]
    \centering
    \includegraphics[scale=0.7]{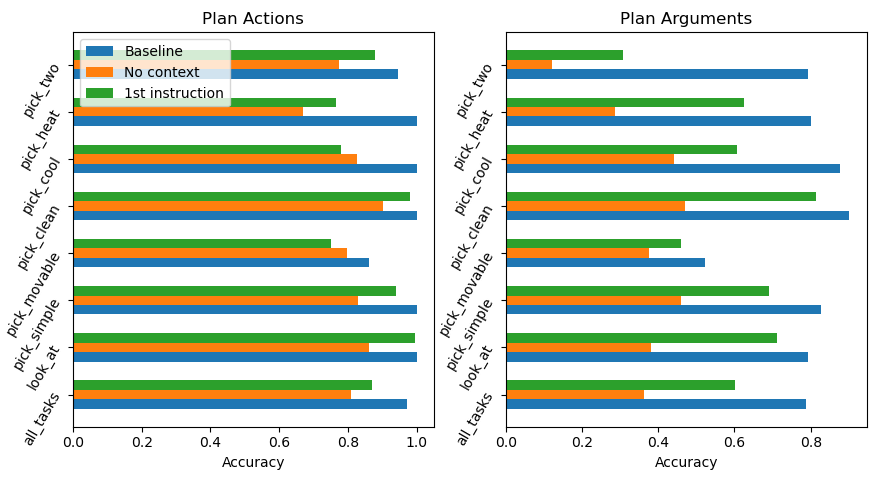}
    \caption{Prediction accuracy of actions and arguments for unseen tasks of a model with a refined context. This experiment compares a model fine-tuned to the task goal and a context consisting of a \gls{NL} description of the first plan instruction (green) with the baseline (blue) and previous RobLM `No context' model (orange).}
    \label{fig:accuracy_first}    
\end{figure}
In a deeper analysis of RobLM failure cases, it has been found that the first argument in the generated plan is the hardest to predict correctly by the \gls{lm}. The \gls{lm} is not able to draw enough conclusions about the first instruction from the supplied context of any form.

This causality becomes obvious after the following experiment: Given the task goal and a \gls{NL} \textbf{description} of the first instruction as context, how does the overall accuracy of the \gls{lm} change? The following text is an example of an instruction description in \gls{NL}, as found in the ALFRED dataset:
\begin{verbatim}
Turn left and walk across the room towards the shelves on the wall.
\end{verbatim}

The results in Figure \ref{fig:accuracy_first} show that, given this extra information, RobLM is almost able to reach the performance levels of the baseline measurement across all tasks; it shows very high accuracy on ``full plan'' actions and arguments. The conclusion of this experiment is that the more precisely the supplied context is tailored towards the key issue of \gls{lm} generation task, the more accurate the generated plan becomes. For this specific problem, finding the correct first argument is key to a successful plan, and with a \gls{NL} description of the first instruction, the \gls{lm} is able to draw the necessary connections from context to plan.
        
The overall conclusion of this observation is that \gls{lm} are adaptive. In other words: \gls{lm} are able to adapt new information into the plan generation, towards a more accurate sequence of instructions.

\subsection{Run-time analysis}
\begin{table}
    \caption{Comparison of inference speeds --- RobLM  against baseline. GPU acceleration used for \gls{lm} (NVIDIA® GeForce RTX 2080 SUPER). The timer starts only after the program or model has been loaded into memory, i.e., only computation (inference) time is measured. `No context' has a maximum token sequence length of 200 and `Full context' has a maximum length of 1024 tokens for generation.}
    \label{tab:times}
    \centering
    \begin{tabular}{c|c|c}
        \hline
        \multicolumn{3}{c}{Iterations per second (average over 800 samples)} \\
        \hline
        Baseline & RobLM `No context' & RobLM `Full context'\\
        \hline
        2.9 & 1.0 & 0.2 \\
        \hline
    \end{tabular}

\end{table}
Time is an important factor when it comes to real-life applications. This is especially true for industrial robotics, where cycle times are important. But not every robotic application is time-critical, e.g., a household robot is not expected to respond in a sub-second time. However, if task planning is seen as a programming problem, a fast execution time greatly enhances the operator experience.

Table \ref{tab:times} shows a comparison of the inference speeds of RobLM against the baseline (\gls{fd}). RobLM, in all cases, is very slow compared to the baseline, which is likely due to the reliance on the full GPT-2 vocabulary size for the \gls{lm} tokenizer and the usage of a \gls{lm}-internal, implementation-specific generation function\footnote{Huggingface generation function: `transformers/src/transformers/generation\_utils.py'}.

\section{Conclusion}\label{sec:conclusion}
We presented a framework for finetuning grounded Large Language Models (LLMs) and investigated the applicability of such models combined with planning in solving ling-horizon robot reasoning tasks. This paper has shown that LLMs can extract commonsense knowledge through precise queries and adjust their behavior based on available information or context. Among our contributions are the development of RobLM, a grounded finetuned LLM that generates plans directly from natural language commands, and Graph2NL, which creates natural language text describing graph-based data, to represent scene graphs as inputs into RobLM. Our extensive experimental results have revealed, nevertheless, the challenges in representing structured and geometric data in natural language.

However, LLMs still need to demonstrate a consistent ability to perform long-horizon planning tasks and, therefore, cannot replace classical planning systems. Despite their limitations, LLMs possess powerful features such as efficient storage and retrieval of commonsense knowledge, which can be useful in planning tasks when presented with partially observable environments.

For future work, exploring larger models like GPT-3 or GPT-NeoX could increase the accuracy and success rate of RobLM. Providing structured context to the Transformer model and exploring multi-modal inputs, such as visual information, may also improve the planning capabilities of LLMs. Further research in the field of applied natural language processing in robotics could help unlock the full potential of LLMs and contribute to the development of more advanced neuro-symbolic planning systems.

\section*{Conflict of Interest Statement}
The authors declare that the research was conducted in the absence of any commercial or financial relationships that could be construed as a potential conflict of interest.



\section*{Funding}
This research has been supported by the German Research Foundation (DFG) through the Emmy Noether Programme (CH 2676/1-1) and by the Hessian.AI Connectom Fund ``Robot Learning of Long-Horizon Manipulation bridging Object-centric
Representations to Knowledge Graphs''.

\section*{Acknowledgments}
The authors would like to acknowledge Snehal Jauhri and Haau-Sing Li for the fruitful discussions and suggestions.


\section*{Data Availability Statement}
Code of the presented research is publicly available through the link: \url{https://github.com/dnandha/RobLM.git}

\bibliographystyle{unsrt}  

\bibliography{references}

\end{document}

%% file: acronyms.tex
 \newacronym{adl}{ADL}{Action Description Language}
 \newacronym{ai}{AI} {Artificial Intelligence}
 
 \newacronym{bc}{BC}{Behavioral Cloning}
 \newacronym{bpe}{BPE}{Byte-Pair Encoding}
 
 \newacronym{clm}{CLM}{Causal Language Modeling}
 \newacronym{cnn}{CNN}{Convolutional Neural Network}
 
 \newacronym{fd}{FD}{Fast Downward}
 \newacronym{fsl}{FSL}{Few-Shot Learning}
 \newacronym{gnn}{GNN}{Graph Neural Network}
 
 \newacronym{il}{IL}{Imitation Learning}
 \newacronym{irl}{IRL}{Inverse Reinforcement Learning}
 
 \newacronym{kb}{KB}{Knowledge Base}
 \newacronym{kg}{KG}{Knowledge Graph}
 
 \newacronym{lm}{LM}{Language Model}
 \newacronym{llm}{LLMs} {Large Language Models}
 
 \newacronym{mdp}{MDP}{Markov Decision Process}
 \newacronym{ml}{ML}{Machine Learning}
 \newacronym{mle}{MLE}{Maximum Likelihood Estimator}
 \newacronym{mlm}{MLM}{Masked Language Modeling}
 
 \newacronym{NL}{NL}{Natural Language}
 \newacronym{nlp}{NLP}{Natural Language Processing}
 \newacronym{NN}{NN}{Neural Network}

 \newacronym{PDDL}{PDDL}{Planning Domain Description Language}
 \newacronym{plm}{PLM} {Pre-trained Language Model}
 \newacronym{pomdp}{POMDP}{Partially Observable Markov Decision Process}
 
 \newacronym{rl}{RL}{Reinforcement Learning}
 \newacronym{rnn}{RNN}{Recurrent Neural Network}
 
 \newacronym{Seq2Seq}{Seq2Seq}{Sequence To Sequence}
 \newacronym{STRIPS}{STRIPS}{Stanford Research Institute Problem Solver}

 \newacronym{TAMP}{TAMP}{Task And Motion Planning}

 \newacronym{ZSL}{ZSL}{Zero-Shot Learning}

\newacronym{roblm}{RobLM}{Robot Language Model}

\newglossary[opg]{operator}{opi}{opo}{List of Operators}
\newglossaryentry{ln}{
  sort        = {natural logarithm},
  name        = {\ensuremath{\mathrm{ln}}},
  symbol      = {\ensuremath{\mathrm{ln}\left(\; \bullet \;\right)} },
  description = {the natural logarithm},
  type        = operator
}

\makeglossaries